\documentclass[sigconf,nonacm]{acmart}

\usepackage{algorithm}
\usepackage{algpseudocode}
\usepackage{hyperref}

\AtBeginDocument{%
  }

\begin{document}

\title{DyNA-Mixer: Decoupling Time and Topology for Scalable Dynamic Node Affinity Prediction}


\author{Yuanxing Zheng}
\affiliation{%
  \institution{East China Normal University}
  \city{Shanghai}
  \country{China}}
\email{51285902051@stu.ecnu.edu.cn}

\author{Liping Wang}
\correspondingauthor
\affiliation{%
  \institution{East China Normal University}
  \city{Shanghai}
  \country{China}}
\email{lipingwang@sei.ecnu.edu.cn}

\author{Xuemin Lin}
\affiliation{%
  \institution{Shanghai Jiao Tong University}
  \city{Shanghai}
  \country{China}}
\email{xuemin.lin@gmail.com}







\renewcommand{\shortauthors}{Zheng et al.}

\begin{abstract}

Continuous-time dynamic graphs preserve the ordering, timing, and evolving strength of interactions that static snapshots discard, making dynamic graph mining essential for evolving systems such as recommender systems, trade networks, and transaction platforms. Research in this area has centered on dynamic link prediction, which forecasts only whether a future interaction will occur, whereas many applications also need its intensity. The dynamic node affinity prediction (DNAP) problem generalizes this task to continuous targets and full-candidate ranking, making it more broadly applicable but substantially harder. This raises two coupled difficulties: ranking requires a predictor expressive enough to capture both multi-scale temporal evolution and graph-wide collaborative patterns, yet running such a predictor over every source--destination pair is computationally prohibitive. We propose \textbf{DyNA-Mixer}, which decouples temporal forecasting from collaborative structure modeling. It adopts a multi-scale temporal predictor for pair-specific affinity evolution, constructs a Virtual Collaborative Graph that infers query-time relationships and aggregates trends from related interactions, and fuses the two through a gated residual mechanism. A sparsity-aware evaluation strategy further restricts expensive forecasting to pairs with recent non-zero affinities. Experiments on four Temporal Graph Benchmark datasets for the DNAP problem show that DyNA-Mixer achieves new state-of-the-art NDCG@10 on every dataset, including a 17.8\% relative test improvement over the strongest baseline on tgbn-token, the largest dataset.
\end{abstract}





\received{20 February 2026}
\received[revised]{12 March 2026}
\received[accepted]{5 June 2026}

\maketitle

\section{Introduction}

Continuous-time dynamic graphs represent relational systems as a chronological
stream of timestamped, and often weighted, interactions among entities
\cite{huang2023temporal}. Compared with static graphs, which collapse an entire interaction history into a single aggregated topology, dynamic graphs better preserve the ordering, timing, and evolving strength of interactions, as well as the continual emergence and disappearance of entities. These properties have made dynamic graph mining the standard tool for intrinsically evolving systems, including recommender systems, social platforms, trade networks, and financial transaction analysis.


Dynamic link prediction has become the dominant task in dynamic graph mining and
the de facto benchmark for representation quality: given a node pair and a query
time, the model forecasts whether the two nodes will interact. Research has
centered on Temporal Graph Neural Networks (TGNNs), which fall into two families:
memory-based methods that maintain a continuously updated state per node
\cite{trivedi2018dyrep,DBLP:journals/corr/abs-2006-10637}, and
neighbor-aggregation methods that encode temporally ordered historical neighbors
at query time \cite{Xu2020Inductive,yu2023towards}. Recent studies have further
questioned this architectural complexity, showing that lightweight encoders over
recent link histories remain competitive \cite{cong2023do}, and have tightened
evaluation with harder negative sampling \cite{huang2023temporal}. 

\begin{figure*}[t]
  \centering
  \includegraphics[width=\textwidth]{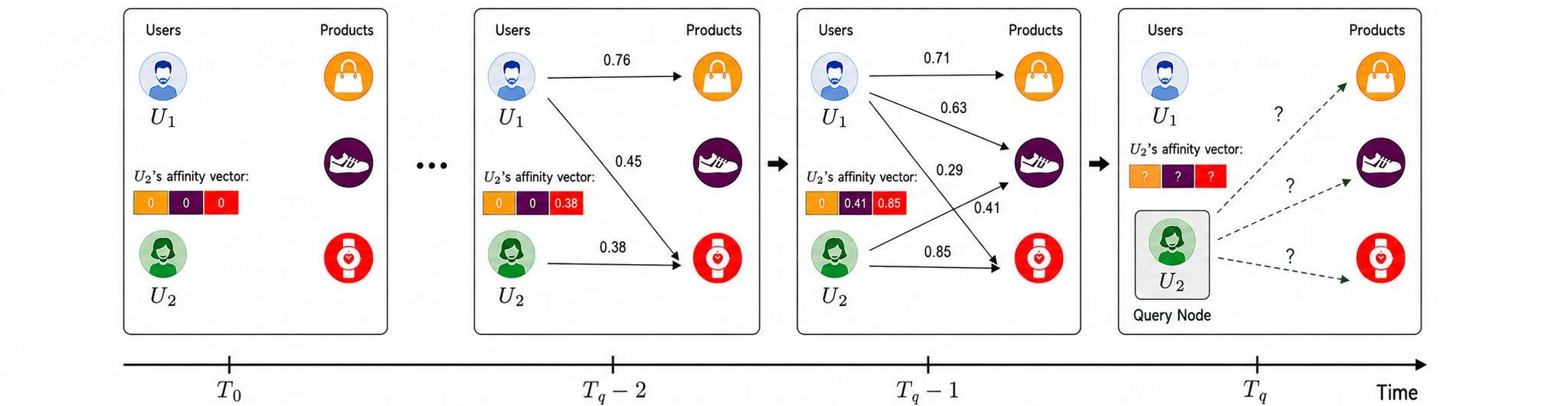}
  \caption{\textbf{Illustration of the dynamic node affinity prediction task.} Unlike traditional dynamic link prediction that infers the binary existence of specific edges, affinity prediction models the continuous evolution of interaction weights over time. At the prediction query time, the model must forecast the continuous affinity scores from the query node (e.g., $U_2$) to all candidate destinations.}
  \Description{A user-product dynamic bipartite graph observed over time. Historical weighted interactions determine the evolving affinity vector of user U2. At the query time, the task is to predict U2's future affinity with every candidate product.}
  \label{fig:teaser}
\end{figure*}

Dynamic link prediction, however, focuses on a binary decision that indicates only
whether an interaction occurs, not how strong it will be. This limits its use in
applications that are inherently quantitative and comparative: a recommender must
order an entire catalog, and trade or financial forecasting concerns interaction
volumes rather than mere occurrence.
The Temporal Graph Benchmark (TGB) formalizes this requirement as the
\textit{dynamic node affinity prediction} (DNAP) problem \cite{huang2023temporal}.
Given a query node $u$ and a query time $T_q$, and conditioned on all observations
before $T_q$, the model must predict $u$'s continuous affinity toward \emph{every}
candidate destination and rank the destinations by the predicted scores, as
illustrated on a user--product graph in Figure~\ref{fig:teaser}.
DNAP is therefore a more broadly applicable formulation than dynamic link prediction, but also a substantially harder one, as accurate ranking requires calibrated intensity estimates over the entire destination space.


\noindent\textbf{Challenges.} The DNAP problem presents two main challenges. First, a model must capture both the temporal evolution of each source-destination pair and collaborative patterns shared across the graph. Existing TGNNs often underperform simple historical heuristics on this task \cite{huang2023temporal,cong2023do,tjandra2025enhancing}. NAVIS \cite{feldman2026revisting} improves upon these heuristics through linear State Space Models (SSMs), but its linear sequence formulation has limited capacity to represent temporal patterns at different scales, while its global virtual state does not explicitly model localized collaborative relationships. Second, employing a more expressive temporal predictor at scale is non-trivial. A straightforward implementation would apply the predictor independently to every source-destination pair. Its cost increases with the number of source nodes to be evaluated, the number of destination classes considered for each source, and the length of the historical sequence processed for every pair. As these three quantities grow, dense evaluation requires both more sequences and more computation per sequence, resulting in substantial runtime and memory costs.

\noindent\textbf{Our Solution.} To address these challenges, we propose DyNA-Mixer, a framework that decouples temporal forecasting from collaborative structure modeling. DyNA-Mixer uses multi-scale temporal forecasting to capture the evolution of each pair and constructs a Virtual Collaborative Graph (VCG) at query time to aggregate relevant trends from related interactions. A gated residual mechanism then combines the pair's own history with this collaborative context. To make the expressive temporal predictor practical over the full candidate space, we design a Sparse-Aware Evaluation Strategy that bypasses deep computation for pairs with empty historical windows. Consequently, expensive forecasting is performed only for active pairs instead of being naively repeated across the complete pairwise space.

Our main contributions are summarized as follows:
\begin{itemize}
    \item We propose DyNA-Mixer, which combines multi-scale temporal forecasting with Virtual Collaborative Graph aggregation to model both pair-specific temporal evolution and collaborative trends.

    \item We design a Sparse-Aware Evaluation Strategy that makes expressive temporal forecasting practical for the DNAP problem by avoiding unnecessary computation over inactive pairs. We analyze its computational properties and empirically evaluate its efficiency-accuracy trade-off.

    \item Experiments on four Temporal Graph Benchmark datasets \cite{huang2023temporal} show that DyNA-Mixer achieves state-of-the-art performance across all datasets, including a 17.8\% relative improvement over the strongest baseline on the largest dataset.
\end{itemize}
\section{Related Work}

\noindent\textbf{Temporal Graph Learning for Link Prediction.} Representation learning on dynamic graphs has been extensively studied \cite{DBLP:journals/jmlr/KazemiGJKSFP20}. TGNNs typically process interactions either by maintaining recurrent node states, as in JODIE, DyRep, and TGN \cite{DBLP:conf/kdd/KumarZL19,trivedi2018dyrep,DBLP:journals/corr/abs-2006-10637}, or by directly encoding historical temporal neighborhoods. The latter family includes attention-based aggregation in TGAT \cite{Xu2020Inductive}, causal anonymous walks in CAWN \cite{DBLP:conf/iclr/WangCLL021}, contrastive temporal modeling in TCL \cite{DBLP:journals/corr/abs-2105-07944}, mixer-based architectures in GraphMixer \cite{cong2023do}, and neighbor co-occurrence encoding in DyGFormer \cite{yu2023towards}. Recent approaches further improve robustness and scalability by filtering sequential dynamics, unifying structural evolution across temporal granularities, or redesigning temporal message passing for large graphs \cite{10.1145/3774904.3792101,DBLP:journals/tkde/XuZLZ25,DBLP:conf/icde/000200O024}. These methods have demonstrated strong performance on binary temporal link prediction, but their objectives and inductive biases do not directly address ranking continuous affinities over the full destination set.

\noindent\textbf{Node Affinity Prediction.} Earlier studies modeled interaction strengths through weighted temporal link prediction \cite{DBLP:conf/infocom/LeiQ0ZY19,DBLP:journals/tkde/QinZBZY23,DBLP:journals/tcyb/YangLCZCS20}. Ranking-oriented methods such as BandRank directly optimize future candidate ordering with frequency-disentangled temporal representations and a dedicated ranking objective \cite{li2025ranking}. The Temporal Graph Benchmark (TGB) \cite{huang2023temporal} introduced the DNAP problem, which requires forecasting continuous interaction intensities over the full destination set. Its evaluation showed that complex TGNNs are often outperformed by simple heuristics, such as moving averages, on this task. To address this, TGNv2 \cite{tjandra2025enhancing} augments standard TGNNs with source-target identifiers, enabling the model to represent moving averages by overcoming the inherent permutation invariance of graph aggregators. Building upon these insights, NAVIS \cite{feldman2026revisting} further generalizes these heuristic properties through discrete linear SSMs. It also employs LambdaRank-based optimization \cite{DBLP:conf/nips/BurgesRL06,DBLP:journals/jmlr/BurgesSBPW11,DBLP:conf/cikm/WangLGBN18} to directly target ranking quality. While these sequence-centric models have achieved state-of-the-art results, the explicit integration of localized collaborative dependencies remains an open direction.

\noindent\textbf{Temporal Sequence Modeling.} Modern sequence models increasingly target long-range dependencies with efficient structured state spaces, including S4 \cite{DBLP:conf/nips/GuJGSDRR21,DBLP:conf/iclr/GuGR22} and Mamba \cite{DBLP:journals/corr/abs-2312-00752}; related formulations have also been adapted to continuous-time dynamic graphs \cite{ding2025dygmamba,li2025dygmamba}. Modeling different temporal horizons is also useful in recommendation, where long-term and short-term user interests can be encoded separately to capture evolving preferences \cite{DBLP:conf/aaai/OuZZLLA25}. In time-series forecasting, TimeMixer \cite{wang2024timemixer} exploits the distinct behaviors exhibited by real-world sequences at different sampling scales. Its Past-Decomposable-Mixing blocks construct multi-scale histories and separate macroscopic trends from microscopic seasonal fluctuations. Treating each pair's historical affinity as a time series enables this multi-scale design to capture temporal inertia, while our framework further complements it with localized collaborative signals from the evolving graph.
\section{Preliminaries}

In this section, we introduce continuous-time dynamic graphs, formally define the DNAP problem, and describe its ranking-based evaluation metric.

\subsection{Continuous-Time Dynamic Graphs (CTDGs)}
We represent a continuous-time dynamic graph (CTDG) as a chronological stream of timestamped interaction events. To accommodate diverse graph structures, such as homogeneous trade networks and user-item bipartite graphs, we define $\mathcal{V}_{src}$ as the set of source nodes and $\mathcal{V}_{tgt}$ as the set of destination nodes. For homogeneous graphs, $\mathcal{V}_{src} \equiv \mathcal{V}_{tgt}$.

The dynamic graph observed up to time $t$ is represented as
\begin{equation}
\mathcal{G}_t = \{(u_j,v_j,\tau_j,\mathbf{e}_j)\}_{j=1}^{J(t)},
\end{equation}
where each tuple denotes an interaction from a source node $u_j \in \mathcal{V}_{src}$ to a destination node $v_j \in \mathcal{V}_{tgt}$ at time $\tau_j \in \mathbb{R}^{+}$, with $\tau_j \leq t$. The vector $\mathbf{e}_j \in \mathbb{R}^{d_e}$ contains the interaction attributes, such as its weight or intensity, and $J(t)$ is the number of interactions observed by time $t$.

\subsection{Dynamic Node Affinity Prediction}
\label{sec:dnap_definition}
Let $N=|\mathcal{V}_{src}|$ and $C=|\mathcal{V}_{tgt}|$ denote the numbers of source and destination nodes, respectively. Given a query source node $u \in \mathcal{V}_{src}$ at time $t$, the DNAP problem requires forecasting its interaction intensity with every destination node during a future interval $[t,t+\delta]$. Let $\mathbf{y}_u(t) \in \mathbb{R}^{C}$ denote the ground-truth affinity vector, where $\mathbf{y}_u(t)[v]$ is the normalized aggregate interaction intensity from $u$ to destination $v \in \mathcal{V}_{tgt}$ during this interval.

The objective is to learn a parameterized model $\mathcal{F}_{\theta}$ that maps the query node and the graph history to a predicted affinity score vector:
\begin{equation}
\hat{\mathbf{s}}_u(t)=\mathcal{F}_{\theta}(u,\mathcal{G}_t)\in\mathbb{R}^{C}.
\end{equation}
The scores in $\hat{\mathbf{s}}_u(t)$ induce a ranking over the complete destination set, which is evaluated against the future affinities in $\mathbf{y}_u(t)$.

\subsection{Evaluation Metric}
Performance on the DNAP problem is evaluated by how accurately the predicted scores rank destinations according to their future affinities. To measure this ranking quality, we use Normalized Discounted Cumulative Gain (NDCG) \cite{DBLP:journals/tois/JarvelinK02}, which assigns greater importance to relevant destinations appearing near the top of the predicted ranking.

Let $\pi$ be the permutation obtained by sorting $\hat{\mathbf{s}}_u(t)$ in descending order, such that $\pi(i)$ is the destination ranked at position $i$. The Discounted Cumulative Gain is
\begin{equation}
\operatorname{DCG}(\pi)
=\sum_{i=1}^{C}\frac{2^{\mathbf{y}_u(t)[\pi(i)]}-1}{\log_2(i+1)}.
\end{equation}
Let $\pi^{*}$ denote the ideal permutation obtained by sorting the ground-truth affinities in descending order. NDCG is then defined as
\begin{equation}
\operatorname{NDCG}(\pi)
=\frac{\operatorname{DCG}(\pi)}{\operatorname{DCG}(\pi^{*})}.
\end{equation}
This normalization produces a score of one for an ideal ranking and enables comparison across query nodes with different affinity distributions.
\section{Methodology}

The DNAP problem requires ranking all destinations for a query source from pre-query interactions. DyNA-Mixer combines evolving graph context with each candidate pair's affinity history. At time $t$, its inputs are the dynamic edge stream $\mathcal{G}_{t}$, whose events update graph states, and each pair's length-$L$ history, constructed from the ground-truth vectors in Section~\ref{sec:dnap_definition}. Let $t_1<\cdots<t_q$ be consecutive label timestamps, with $t_q$ the current query. For pair $(u,c)$, we collect destination $c$'s affinity from the preceding $L$ vectors:

\begin{equation}
H_{u,c}(t_q)
=
\big[
\mathbf{y}_{u}(t_{q-L})[c],\ldots,
\mathbf{y}_{u}(t_{q-1})[c]
\big]
\in\mathbb{R}^{L}.
\end{equation}

Entries run from oldest to newest and are left-padded with zeros when fewer than $L$ labels exist. The prediction target $\mathbf{y}_{u}(t_q)$ is excluded. We write $H_{u,c}$ when the query time is clear.

\begin{figure*}[t]
  \centering
  \includegraphics[width=0.92\textwidth]{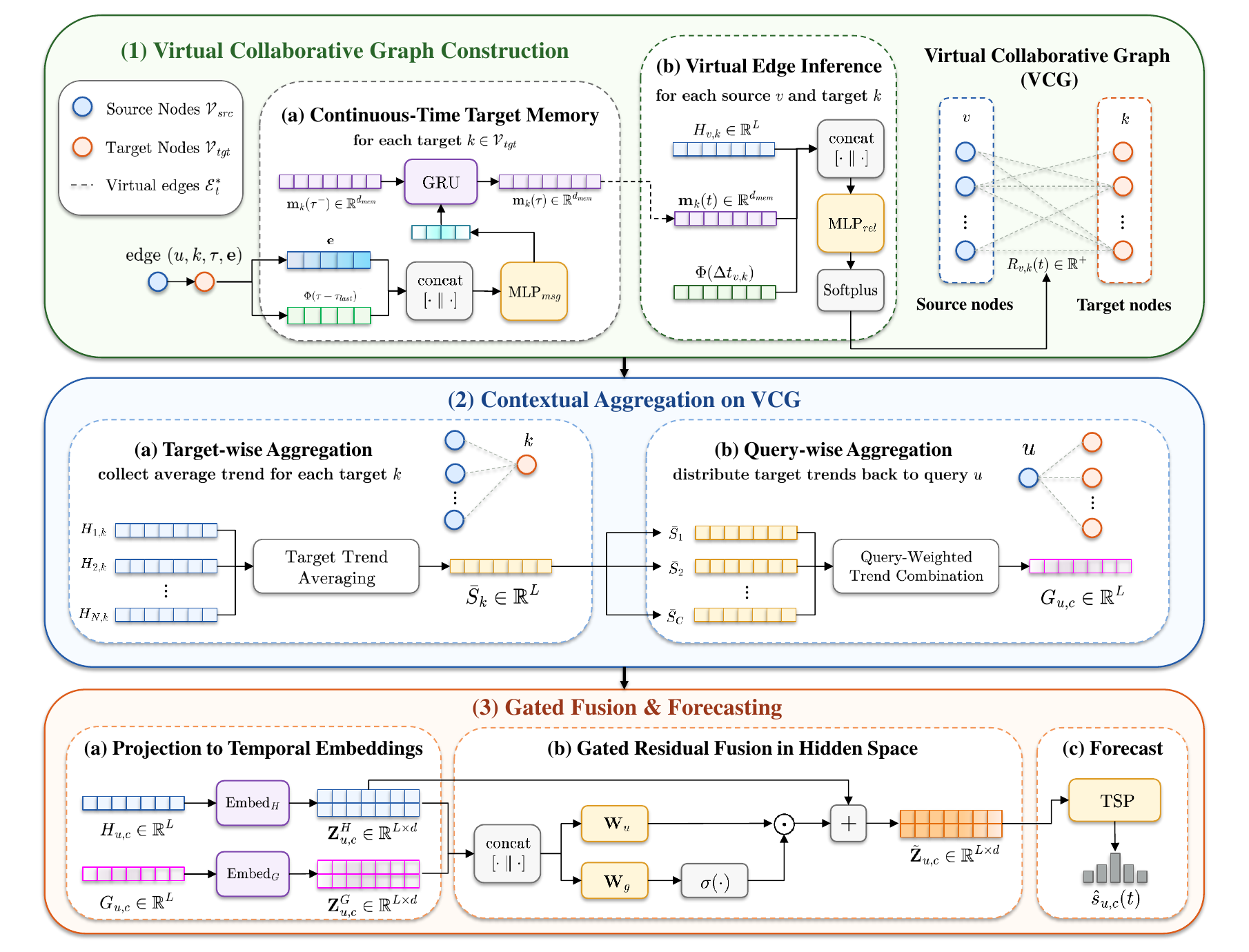}
  \caption{Overview of DyNA-Mixer. The framework consists of three stages: (1) constructing the Virtual Collaborative Graph from continuous-time target memories and inferred virtual edges; (2) aggregating target trends and combining them for the query node; and (3) embedding and fusing self-history with graph-history for temporal forecasting.}
  \Description{The DyNA-Mixer framework constructs a virtual collaborative graph, performs target-wise and query-wise contextual aggregation, and fuses the resulting graph-history with self-history to forecast affinity scores.}
  \label{fig:framework}
\end{figure*}

As illustrated in Figure~\ref{fig:framework}, DyNA-Mixer operates in three stages. First, it constructs a Virtual Collaborative Graph (VCG) to model dynamic source--target relationships. Second, we design a two-stage aggregation mechanism that transforms target trends into query-specific graph context. Third, it fuses this context with pair history for temporal forecasting. To reduce the resulting time and memory costs over the complete candidate space, DyNA-Mixer further exploits temporal sparsity to skip unnecessary computation for inactive pairs.

\subsection{Virtual Collaborative Graph Construction}
\label{sec:vcg_construction}

To capture collaboration among sources interacting with related targets, we construct a query-time VCG $\mathcal{G}^{*}_{t}=(\mathcal{V}_{src},\mathcal{V}_{tgt},\mathcal{E}^{*}_{t})$. Each virtual edge carries a dynamic strength inferred from interaction history by maintaining continuous-time target states and using them to infer edge weights.

\noindent\textbf{Continuous-Time Target Memory.} To track changing target relevance, we maintain a continuous-time memory $\mathbf{m}_{k}(t)\in\mathbb{R}^{d_{mem}}$ that summarizes each target $k$ up to time $t$. When an event $(u,k,\tau, \mathbf{e})$ occurs, we update this memory using a GRU:

\begin{equation}
\mathbf{m}_k(\tau) = \text{GRU} \Big( \mathbf{m}_k(\tau^-), \, \text{MLP}_{msg}\big( [\mathbf{e} \parallel \Phi(\tau - \tau_{last})] \big) \Big),
\end{equation}

Here, $\tau_{last}$ is the latest interaction time of target $k$, and $\Phi(\cdot)$ maps elapsed time to a continuous representation.

\noindent\textbf{Virtual Edge Inference.} Aggregation needs both global evolution and each pair's relevance at the query time. We infer the non-negative weight $R_{v,k}(t)$ from pair history $H_{v,k}$, target memory $\mathbf{m}_{k}(t)$, and elapsed time $\Delta t_{v,k}$ since their latest interaction:

\begin{equation}
R_{v,k}(t) = \text{Softplus}\Big( \text{MLP}_{rel}\big( [H_{v,k} \parallel \mathbf{m}_k(t) \parallel \Phi(\Delta t_{v,k})] \big) \Big).
\end{equation}

Softplus keeps $R_{v,k}(t)$ non-negative, which enables weighted contextual aggregation.

\subsection{Contextual Aggregation on VCG}
\label{sec:vcg_aggregation}

DyNA-Mixer transforms VCG edge weights into temporal context in two stages: summarizing each target's affinity trajectories and then combining them back to the query source node.

\noindent\textbf{Target-wise Aggregation.} Sources connected to one target provide complementary observations of its affinity evolution. We design \textit{Target Trend Averaging} to combine their histories according to the virtual-edge weights, producing $\bar{S}_{k}\in\mathbb{R}^{L}$:

\begin{equation}
\bar{S}_k = \frac{\sum_{v \in \mathcal{V}_{src}} R_{v,k}(t) H_{v,k}}{\sum_{v \in \mathcal{V}_{src}} R_{v,k}(t)}.
\end{equation}

$\bar{S}_{k}$ summarizes target $k$'s affinity evolution across connected sources.

\noindent\textbf{Query-wise Aggregation.} Because target trends differ in relevance to each source, we design \textit{Query-Weighted Trend Combination} to aggregate them for query $u$ according to its normalized virtual-edge weights:

\begin{equation}
G_{u,c} = \sum_{k \in \mathcal{V}_{tgt}} \beta_{u,k} \bar{S}_k
\quad \text{where} \quad
\beta_{u,k} = \frac{R_{u,k}(t)}{\sum_{k'} R_{u,k'}(t)}.
\end{equation}

The resulting $G_{u,c}\in\mathbb{R}^{L}$ provides collaborative temporal context for pair $(u,c)$.

\subsection{Gated Fusion and Forecasting}
\label{sec:gated_forecasting}

To incorporate collaborative graph-history while preserving the information in each pair's own history, the final stage combines $H_{u,c}$ and $G_{u,c}$ in the predictor's hidden space. Their temporal embeddings are defined as follows.

\begin{equation}
\mathbf{Z}^{H}_{u,c} = \mathrm{Embed}_{H}(H_{u,c}), \qquad
\mathbf{Z}^{G}_{u,c} = \mathrm{Embed}_{G}(G_{u,c}).
\end{equation}

Both embeddings lie in $\mathbb{R}^{L\times d}$ for predictor hidden dimension $d$. Gated residual fusion is defined as follows.

\begin{equation}
\tilde{\mathbf{Z}}_{u,c}
=
\mathbf{Z}^{H}_{u,c}
+
\sigma\Big(
\mathbf{W}_{g}
[
\mathbf{Z}^{H}_{u,c}
\parallel
\mathbf{Z}^{G}_{u,c}
]
\Big)
\odot
\mathbf{W}_{u}
[
\mathbf{Z}^{H}_{u,c}
\parallel
\mathbf{Z}^{G}_{u,c}
].
\end{equation}

Here, $[\mathbf{Z}^{H}_{u,c}\parallel\mathbf{Z}^{G}_{u,c}]\in\mathbb{R}^{L\times 2d}$ is channel-wise concatenation. Learned projections $\mathbf{W}_{g},\mathbf{W}_{u}:\mathbb{R}^{2d}\rightarrow\mathbb{R}^{d}$ respectively gate and update the representation; they are applied independently at each temporal position and reused across positions and source--target pairs. The sigmoid controls collaborative context, while the residual preserves self-history. The remaining predictor layers produce the affinity score as follows.

\begin{equation}
\hat{s}_{u,c}(t) = \mathrm{TSP}(\tilde{\mathbf{Z}}_{u,c}).
\end{equation}

We instantiate the predictor with TimeMixer \cite{wang2024timemixer}, whose multi-scale architecture captures different temporal resolutions. Because minimizing errors in individual affinity values does not directly optimize the destination ranking required by the DNAP problem, we follow NAVIS \cite{feldman2026revisting} and train $\hat{\mathbf{s}}_{u}(t)$ with LambdaRank \cite{DBLP:conf/nips/BurgesRL06,DBLP:conf/cikm/WangLGBN18}. The correctly ordered destination pairs for query $u$ are defined as follows.

\begin{equation}
\mathcal{O}_{u}(t)
=
\big\{(i,j)\mid \mathbf{y}_{u}(t)[i]>\mathbf{y}_{u}(t)[j]\big\}.
\end{equation}

Each $(i,j)\in\mathcal{O}_{u}(t)$ requires $i$ above $j$. Its logistic loss is defined as follows.

\begin{equation}
\ell_{ij}
=\log\!\left(1+\exp\!\left[
\hat{s}_{u,j}(t)-\hat{s}_{u,i}(t)\right]\right).
\end{equation}

The NDCG-weighted ranking loss is defined as follows.

\begin{equation}
\mathcal{L}_{\mathrm{rank}}
=\frac{1}{|\mathcal{O}_{u}(t)|}
\sum_{(i,j)\in\mathcal{O}_{u}(t)}
\left|\Delta\mathrm{NDCG}_{ij}\right|\ell_{ij}.
\end{equation}

Here, $|\Delta\mathrm{NDCG}_{ij}|$ is the absolute NDCG change from exchanging $i$ and $j$, assigning larger penalties to misorderings with greater ranking impact.

\subsection{Sparse-Aware Evaluation Strategy}
\label{sec:sparse_evaluation}

Applying this expressive pipeline to every candidate pair is expensive: a naive implementation processes one length-$L$ sequence per pair. Let $\mathcal{Q}_{t}\subseteq\mathcal{V}_{src}$ contain the $B_t$ sources queried at $t$, and let $\mathcal{T}(L,d)$ be the predictor cost for one dimension-$d$ sequence. Dense evaluation has the following complexity.

\begin{equation}
\mathcal{O}\!\left(B_tC\left(Ld^{2}+\mathcal{T}(L,d)\right)\right).
\end{equation}

Our \textit{Sparse-Aware Evaluation Strategy} applies the complete pipeline only to active queried pairs $\mathcal{P}^{L}_{active}(t;\mathcal{Q}_t)$, whose length-$L$ history contains a non-zero affinity; the complement is $\mathcal{P}^{L}_{inactive}(t;\mathcal{Q}_t)$.

To avoid rescanning each length-$L$ history, a cache $\mathbf{A}$ tracks its non-zero count. Each entry at $t_q$ is defined as follows.

\begin{equation}
\mathbf{A}_{u,c}(t_q)
=
\sum_{r=\max(1,q-L)}^{q-1}
\mathbb{I}\!\left[\mathbf{y}_{u}(t_r)[c]>0\right].
\end{equation}

The cache starts at zero, and the non-zero indicator is defined as follows.

\begin{equation}
z_{u,c}(t_r)=\mathbb{I}\!\left[\mathbf{y}_{u}(t_r)[c]>0\right].
\end{equation}

After prediction, observing $\mathbf{y}_{u}(t_q)$ advances the window by removing the expired indicator and adding the new one, yielding the following update.

\begin{equation}
\begin{aligned}
\mathbf{A}_{u,c}(t_{q+1})
&=\mathbf{A}_{u,c}(t_q)-\mathbb{I}[q>L]z_{u,c}(t_{q-L})\\
&\quad+z_{u,c}(t_q).
\end{aligned}
\end{equation}

Adding the target only after prediction prevents leakage. A vectorized comparison of cached counts for $\mathcal{Q}_{t}$ identifies both sets. Algorithm~\ref{alg:sparse_evaluation} assigns $-\infty$ to inactive pairs, predicts active pairs, and then updates the cache from observed labels.

\begin{algorithm}[t]
\caption{Sparse-Aware Evaluation Strategy}
\label{alg:sparse_evaluation}
\begin{algorithmic}[1]
\Statex \hspace{-\algorithmicindent}\textbf{Input:} $\mathcal{G}_{t_q}$, $\mathcal{Q}_{t_q}$, $\mathcal{H}=\{H_{u,c}\}$, $\mathbf{A}(t_q)$
\Statex \hspace{-\algorithmicindent}\textbf{Output:} Predicted scores $\{\hat{s}_{u,c}(t_q)\}$, updated cache $\mathbf{A}(t_{q+1})$
\If{$q=1$}
    \State $\mathbf{A}(t_q)\gets\mathbf{0}$
\EndIf
\State $\mathcal{P}^{L}_{active}(t_q;\mathcal{Q}_{t_q})\gets\{(u,c)\in\mathcal{Q}_{t_q}\times\mathcal{V}_{tgt}\mid\mathbf{A}_{u,c}(t_q)>0\}$
\State $\mathcal{P}^{L}_{inactive}(t_q;\mathcal{Q}_{t_q})\gets(\mathcal{Q}_{t_q}\times\mathcal{V}_{tgt})\setminus\mathcal{P}^{L}_{active}(t_q;\mathcal{Q}_{t_q})$
\State Assign $-\infty$ to all pairs in $\mathcal{P}^{L}_{inactive}(t_q;\mathcal{Q}_{t_q})$
\State $\mathcal{G}^{*}_{t_q}\gets\operatorname{VCGConstruct}(\mathcal{G}_{t_q},\mathcal{H},\mathcal{P}^{L}_{active}(t_q;\mathcal{Q}_{t_q}))$
\ForAll{$(u,c)\in\mathcal{P}^{L}_{active}(t_q;\mathcal{Q}_{t_q})$}
    \State $G_{u,c}\gets\operatorname{ContextAggregate}(\mathcal{G}^{*}_{t_q},\mathcal{H},u,c)$
    \State $\widetilde{\mathbf{Z}}_{u,c}\gets\operatorname{GatedFuse}(H_{u,c},G_{u,c})$
    \State $\hat{s}_{u,c}(t_q)\gets\operatorname{TSP}(\widetilde{\mathbf{Z}}_{u,c})$
\EndFor
\State \textit{After $\mathbf{y}(t_q)$ is observed:}
\ForAll{$(u,c)\in\mathcal{Q}_{t_q}\times\mathcal{V}_{tgt}$}
    \State $a_{out}\gets\mathbb{I}[q>L\ \land\ \mathbf{y}_{u}(t_{q-L})[c]>0]$
    \State $a_{in}\gets\mathbb{I}[\mathbf{y}_{u}(t_q)[c]>0]$
    \State $\mathbf{A}_{u,c}(t_{q+1})\gets\mathbf{A}_{u,c}(t_q)-a_{out}+a_{in}$
\EndFor
\State \Return $\{\hat{s}_{u,c}(t_q)\}$, $\mathbf{A}(t_{q+1})$
\end{algorithmic}
\end{algorithm}

Identifying support and filling the required $B_t\times C$ output costs $\mathcal{O}(B_tC)$ lightweight operations. Restricting VCG, fusion, and forecasting to $|\mathcal{P}^{L}_{active}(t;\mathcal{Q}_t)|$ pairs gives the following complexity.

\begin{equation}
\mathcal{O}\!\left(
B_tC+|\mathcal{P}^{L}_{active}(t;\mathcal{Q}_t)|
\left(Ld^{2}+\mathcal{T}(L,d)\right)
\right).
\end{equation}

Across timestamps, support costs $\mathcal{O}(C\sum_t B_t)$, reaching $\mathcal{O}(NCT)$ only when every source is always queried. Crucially, inactive pairs avoid the factor $Ld^{2}+\mathcal{T}(L,d)$. Appendix~\ref{app:theoretical_analysis} derives time and memory costs, proves the sparsity bound, and discusses ranking preservation; Section~\ref{sec:empirical_sparsity} measures activity empirically.

\section{Experiments}

\subsection{Experimental Setup}

\begin{table*}[t]
\centering
\caption{Official statistics of the TGB datasets used to evaluate the DNAP problem.}
\label{tab:dataset_statistics}
\resizebox{0.92\textwidth}{!}{
\begin{tabular}{llrrrr}
\toprule
\textbf{Scale} & \textbf{Dataset} & \textbf{Node Count} & \textbf{Temporal Edge Count} & \textbf{Temporal Step Count} & \textbf{Surprise} \\
\midrule
small  & tgbn-trade  & 255    & 468,245    & 32         & 0.023 \\
medium & tgbn-genre  & 1,505  & 17,858,395 & 133,758    & 0.005 \\
large  & tgbn-reddit & 11,766 & 27,174,118 & 21,889,537 & 0.013 \\
large  & tgbn-token  & 61,756 & 72,936,998 & 2,036,524  & 0.014 \\
\bottomrule
\end{tabular}
}
\end{table*}

\noindent\textbf{Datasets.} We evaluate DyNA-Mixer on four TGB datasets for the DNAP problem \cite{huang2023temporal}, whose official statistics appear in Table~\ref{tab:dataset_statistics}; surprise is the fraction of test edges unseen during training. tgbn-trade predicts annual agricultural trade intensity between countries \cite{7b4a0f19-d7f8-3f6d-b970-7e311fca30f4}; tgbn-genre derives evolving music-genre preferences from the Million Song Dataset \cite{DBLP:conf/ismir/Bertin-MahieuxEWL11}; tgbn-reddit models changing user affinities toward Reddit communities \cite{10.1145/3487553.3524699}; and tgbn-token captures account--token affinities from blockchain transactions \cite{shamsi2022chartalist}.

\noindent\textbf{Baselines.} We compare three historical heuristics and eleven learned baselines in three groups. 

\textit{Historical heuristics} use observed affinities: Persistent Forecast repeats the latest vector, Moving Average uses recent values, and Historical Average averages the observed history \cite{huang2023temporal}. 

\textit{General-purpose temporal graph models} learn from evolving interactions: JODIE and DyRep maintain recurrent node states \cite{DBLP:conf/kdd/KumarZL19,trivedi2018dyrep}; TGN combines event-driven memory and temporal message passing \cite{DBLP:journals/corr/abs-2006-10637}; TGAT uses time-aware neighbor attention \cite{Xu2020Inductive}; CAWN encodes causal anonymous walks \cite{DBLP:conf/iclr/WangCLL021}; TCL uses contrastive temporal learning \cite{DBLP:journals/corr/abs-2105-07944}; GraphMixer mixes link encodings and node features \cite{cong2023do}; DyGFormer uses a Transformer with neighbor co-occurrence encodings \cite{yu2023towards}; and DyGMamba applies selective state-space modeling \cite{ding2025dygmamba}. 

\textit{Models designed for the DNAP problem} capture pair-specific affinity histories: TGNv2 adds source--target identities \cite{tjandra2025enhancing}, while NAVIS uses linear state-space models and directly optimizes ranking \cite{feldman2026revisting}.

\noindent\textbf{Settings.} Following TGB, we use official chronological splits and validation/test NDCG@10. NAVIS supplies baselines: Persistent Forecast uses previous labels; all others use full-graph messages. Validation NDCG@10 selects dataset-specific $L$ and the test-reporting epoch. We train 200 epochs on tgbn-trade and 50 otherwise. Each experiment uses one GPU: a V100 for tgbn-trade/genre and an A100 for tgbn-reddit/token.

\subsection{Main Results}

Table~\ref{tab:main_results} reports validation/test NDCG@10 on four TGB datasets for the DNAP problem, using NAVIS scores with previous labels for Persistent Forecast and full-graph messages otherwise.

\begin{table*}[t]
\centering
\caption{NDCG@10 on TGB datasets for the DNAP problem. Higher is better. The symbol $\pm$ denotes the mean and standard deviation over multiple random seeds; deterministic heuristics are reported without deviations. Persistent Forecast uses previous ground-truth labels, while the remaining baseline results use the full-graph-message setting. Boldface and underlining mark the best and second-best results in each column, respectively.}
\label{tab:main_results}
\resizebox{\textwidth}{!}{
\begin{tabular}{lcccccccc}
\toprule
\raisebox{-0.7\normalbaselineskip}{\textbf{Method}}
& \multicolumn{2}{c}{\textbf{tgbn-trade}}
& \multicolumn{2}{c}{\textbf{tgbn-genre}}
& \multicolumn{2}{c}{\textbf{tgbn-reddit}}
& \multicolumn{2}{c}{\textbf{tgbn-token}} \\
\cmidrule(lr){2-3}
\cmidrule(lr){4-5}
\cmidrule(lr){6-7}
\cmidrule(lr){8-9}
& \textbf{Val} & \textbf{Test} & \textbf{Val} & \textbf{Test} & \textbf{Val} & \textbf{Test} & \textbf{Val} & \textbf{Test} \\
\midrule
Persistent Forecast & 0.860 & 0.855 & 0.350 & 0.357 & 0.380 & 0.369 & 0.403 & 0.430 \\
Moving Avg & 0.793 & 0.777 & 0.496 & 0.497 & 0.498 & 0.480 & 0.401 & 0.414 \\
Historical Avg & 0.793 & 0.777 & 0.478 & 0.472 & 0.499 & 0.481 & 0.402 & 0.415 \\
\midrule
JODIE & $0.394 \pm 0.050$ & $0.374 \pm 0.090$ & $0.358 \pm 0.030$ & $0.350 \pm 0.040$ & $0.345 \pm 0.020$ & $0.314 \pm 0.010$ & -- & -- \\
TGAT & $0.395 \pm 0.140$ & $0.375 \pm 0.070$ & $0.360 \pm 0.040$ & $0.352 \pm 0.030$ & $0.345 \pm 0.010$ & $0.314 \pm 0.010$ & -- & -- \\
CAWN & $0.393 \pm 0.070$ & $0.374 \pm 0.090$ & -- & -- & -- & -- & -- & -- \\
TCL & $0.394 \pm 0.110$ & $0.375 \pm 0.090$ & $0.362 \pm 0.040$ & $0.354 \pm 0.020$ & $0.347 \pm 0.010$ & $0.314 \pm 0.010$ & -- & -- \\
GraphMixer & $0.394 \pm 0.170$ & $0.375 \pm 0.110$ & $0.361 \pm 0.040$ & $0.352 \pm 0.030$ & $0.347 \pm 0.010$ & $0.314 \pm 0.010$ & -- & -- \\
DyGFormer & $0.408 \pm 0.580$ & $0.388 \pm 0.640$ & $0.371 \pm 0.060$ & $0.365 \pm 0.200$ & $0.348 \pm 0.020$ & $0.316 \pm 0.010$ & -- & -- \\
DyGMamba & $0.393 \pm 0.001$ & $0.374 \pm 0.001$ & $0.359 \pm 0.001$ & $0.351 \pm 0.001$ & $0.347 \pm 0.000$ & $0.314 \pm 0.000$ & -- & -- \\
DyRep & $0.394 \pm 0.001$ & $0.374 \pm 0.001$ & $0.357 \pm 0.001$ & $0.351 \pm 0.001$ & $0.344 \pm 0.001$ & $0.312 \pm 0.001$ & $0.151 \pm 0.006$ & $0.141 \pm 0.006$ \\
TGN & $0.445 \pm 0.009$ & $0.409 \pm 0.005$ & $0.443 \pm 0.002$ & $0.423 \pm 0.007$ & $0.482 \pm 0.007$ & $0.408 \pm 0.006$ & $0.251 \pm 0.000$ & $0.200 \pm 0.005$ \\
TGNv2 & $0.807 \pm 0.006$ & $0.735 \pm 0.006$ & $0.481 \pm 0.001$ & $0.469 \pm 0.002$ & $0.544 \pm 0.000$ & $0.507 \pm 0.002$ & $0.321 \pm 0.001$ & $0.294 \pm 0.001$ \\
NAVIS & $\underline{0.872 \pm 0.001}$ & $\underline{0.863 \pm 0.001}$ & $\underline{0.512 \pm 0.001}$ & $\underline{0.520 \pm 0.001}$ & $\underline{0.564 \pm 0.001}$ & $\underline{0.552 \pm 0.001}$ & $\underline{0.423 \pm 0.001}$ & $\underline{0.444 \pm 0.001}$ \\
DyNA-Mixer & $\textbf{0.884} \pm \textbf{0.001}$ & $\textbf{0.869} \pm \textbf{0.003}$ & $\textbf{0.534} \pm \textbf{0.000}$ & $\textbf{0.544} \pm \textbf{0.000}$ & $\textbf{0.605} \pm \textbf{0.000}$ & $\textbf{0.590} \pm \textbf{0.000}$ & $\textbf{0.510} \pm \textbf{0.000}$ & $\textbf{0.523} \pm \textbf{0.001}$ \\
\bottomrule
\end{tabular}
}
\end{table*}

DyNA-Mixer achieves the best mean validation and test NDCG@10 on every dataset. Its standard deviations are at most $0.003$ across multiple random seeds, indicating limited sensitivity to random initialization. General-purpose temporal graph models frequently trail historical heuristics, whereas TGNv2 and NAVIS, designed for the DNAP problem, are stronger, highlighting the value of task-specific modeling.

Compared with NAVIS, the strongest baseline, DyNA-Mixer improves test NDCG@10 by $0.7\%$, $4.6\%$, $6.9\%$, and $17.8\%$ on tgbn-trade, tgbn-genre, tgbn-reddit, and tgbn-token, respectively. The gains increase across benchmarks of increasing scale and are largest on tgbn-token, which has the most nodes and temporal edges, suggesting that DyNA-Mixer remains effective as the interaction space expands.

\subsection{Ablation Study}

To test whether localized collaborative topology complements temporal inertia, Table~\ref{tab:ablation} compares DyNA-Mixer with a temporal-only TimeMixer variant that removes VCG aggregation and predicts solely from local pair histories.

\begin{table*}[t]
\centering
\caption{Ablation study of the VCG module on TGB node affinity prediction datasets. Results are measured by NDCG@10 under the same random seed ($seed=1$). Higher is better.}
\label{tab:ablation}
\resizebox{0.9\textwidth}{!}{
\begin{tabular}{lcccccccc}
\toprule
\raisebox{-0.5\normalbaselineskip}{\textbf{Method}}
& \multicolumn{2}{c}{\textbf{tgbn-trade}}
& \multicolumn{2}{c}{\textbf{tgbn-genre}}
& \multicolumn{2}{c}{\textbf{tgbn-reddit}}
& \multicolumn{2}{c}{\textbf{tgbn-token}} \\
\cmidrule(lr){2-3}
\cmidrule(lr){4-5}
\cmidrule(lr){6-7}
\cmidrule(lr){8-9}
& \textbf{Val} & \textbf{Test} & \textbf{Val} & \textbf{Test} & \textbf{Val} & \textbf{Test} & \textbf{Val} & \textbf{Test} \\
\midrule
TimeMixer only & 0.873 & 0.863 & 0.533 & 0.542 & 0.602 & 0.589 & 0.498 & 0.516 \\
DyNA-Mixer (TimeMixer + VCG) & \textbf{0.885} & \textbf{0.871} & \textbf{0.535} & \textbf{0.544} & \textbf{0.605} & \textbf{0.591} & \textbf{0.510} & \textbf{0.524} \\
\bottomrule
\end{tabular}
}
\end{table*}

The temporal-only variant captures historical inertia but treats pair histories independently. DyNA-Mixer consistently improves it by augmenting each local sequence with related nodes' VCG-based evolutionary patterns, showing that VCG contributes localized structural information rather than merely increasing model capacity.

\subsection{Empirical Temporal Sparsity}
\label{sec:empirical_sparsity}

\begin{figure*}[t]
  \centering
  \includegraphics[width=0.85\textwidth]{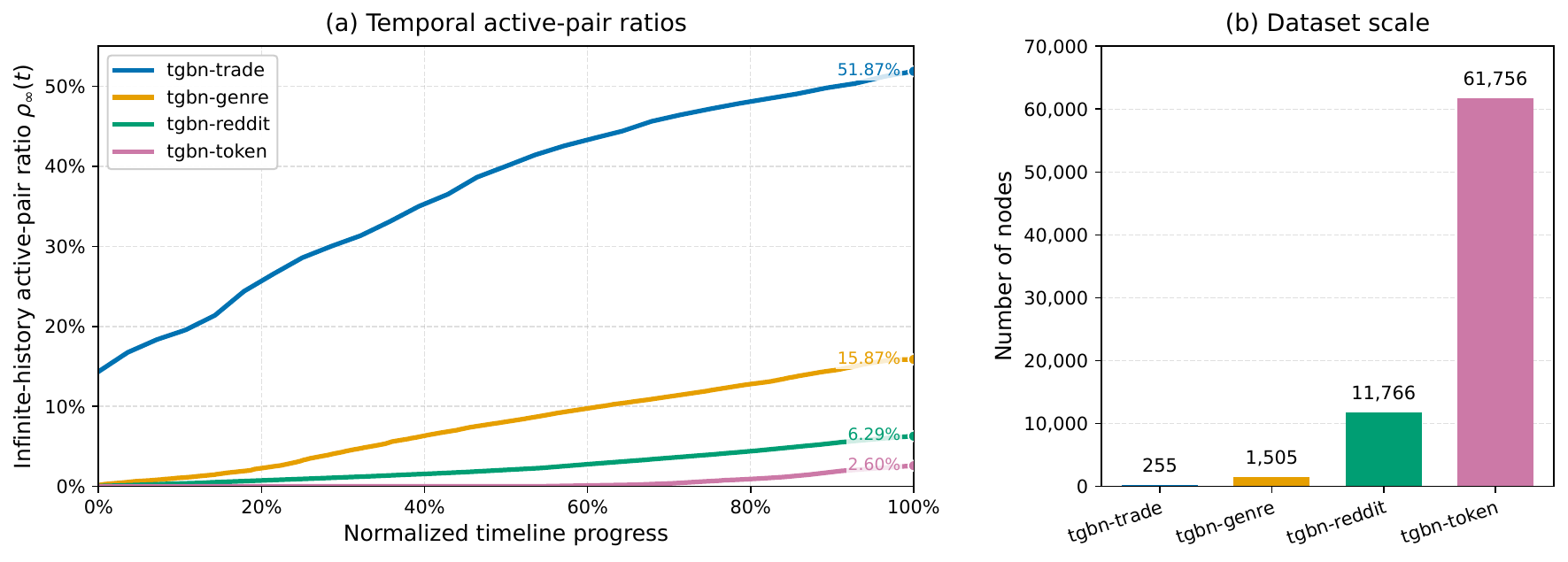}
  \caption{Temporal sparsity and dataset scale. (a) Evolution of the infinite-history active-pair ratio $\rho_{\infty}(t_q)$ over the normalized dataset timeline; endpoint labels report the final ratios. (b) Numbers of nodes in the four datasets, shown on a linear scale. Colors consistently identify datasets across both panels.}
  \label{fig:temporal_sparsity_scale}
\end{figure*}

Sparse-Aware Evaluation depends on how many source--destination pairs can skip forecasting. We quantify this opportunity through each dataset's active-pair ratio. For label timestamps $t_1<\cdots<t_q$ up to query $t_q$ and pairwise space $NC$, the infinite-history active set is

\begin{equation}
\mathcal{P}^{\infty}_{active}(t_q)
=
\left\{(u,c)\in\mathcal{V}_{src}\times\mathcal{V}_{tgt}
\;\middle|\;
\exists r<q,\ \mathbf{y}_u(t_r)[c]>0
\right\}.
\end{equation}

The corresponding cumulative active-pair ratio is

\begin{equation}
\rho_{\infty}(t_q)
=
\frac{\left|\mathcal{P}^{\infty}_{active}(t_q)\right|}{NC}.
\end{equation}

$\rho_{\infty}(t_q)$ is the fraction of pairs with any non-zero affinity before $t_q$. Using complete history makes it independent of the model's finite $L$ and, by Theorem~\ref{thm:infinite_history_bound}, a conservative upper bound on every finite-$L$ active ratio; consequently, $1-\rho_{\infty}(t_q)$ lower-bounds the skippable fraction.

Figure~\ref{fig:temporal_sparsity_scale} shows persistent sparsity even under this conservative infinite-history measure, so every finite history yields an equal or larger inactive fraction. Across the four benchmarks, larger datasets have lower active ratios and hence stronger sparsity. As dense forecasting becomes costlier with $N$, $C$, and $L$, this trend creates more opportunities to skip computation and increases the benefit of Sparse-Aware Evaluation at scale.

\begin{figure*}[t]
  \centering
  \includegraphics[width=\textwidth]{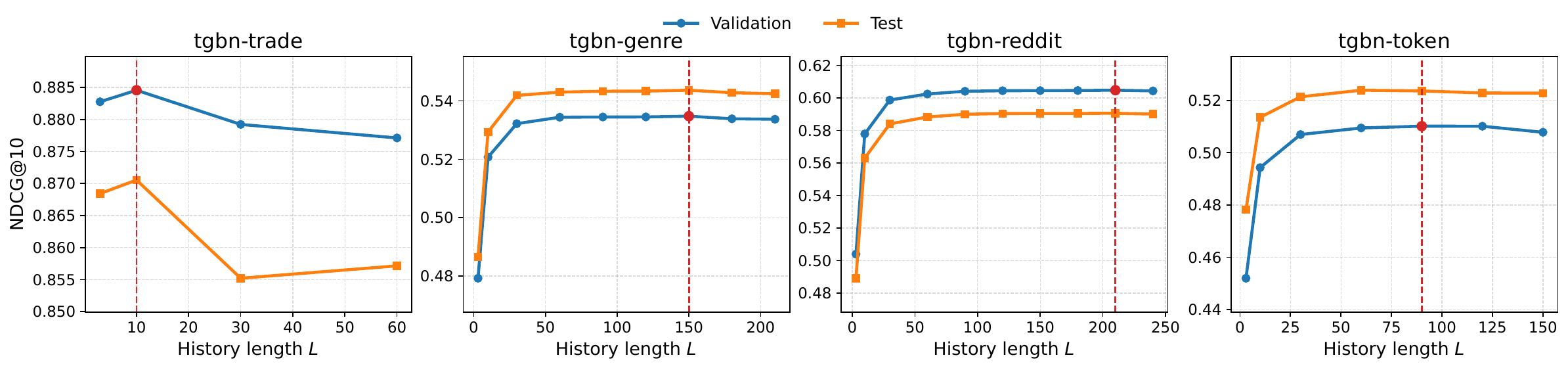}
  \caption{Sensitivity of DyNA-Mixer to the history length $L$. Each panel reports validation and test NDCG@10 under the candidate $L$ values explored for one dataset. The red dashed line marks the history length selected by validation performance.}
  \label{fig:history_length_sensitivity}
\end{figure*}

\begin{table*}[t]
\centering
\caption{Runtime and accuracy comparison between Dense Evaluation and the Sparse-Aware Evaluation Strategy. Runtime is the average training time per epoch. Validation and test results are measured by NDCG@10. OOM indicates that Dense Evaluation exceeds GPU memory capacity.}
\label{tab:runtime_efficiency}
\resizebox{0.92\textwidth}{!}{
\begin{tabular}{lcccccc}
\toprule
\textbf{Dataset} & \textbf{Dense Time} & \textbf{Sparse-Aware Time} & \textbf{Dense Val} & \textbf{Dense Test} & \textbf{Sparse-Aware Val} & \textbf{Sparse-Aware Test} \\
\midrule
tgbn-trade  & 6.4s    & 16.9s   & 0.885 & 0.870 & 0.885 & 0.871 \\
tgbn-genre  & 1057.2s & 726.7s  & 0.535 & 0.543 & 0.535 & 0.543 \\
tgbn-reddit & OOM     & 4721.4s & --    & --    & 0.605 & 0.591 \\
tgbn-token  & OOM     & 3470.8s & --    & --    & 0.510 & 0.524 \\
\bottomrule
\end{tabular}
}
\end{table*}

\subsection{Runtime Efficiency}

We evaluate the efficiency--accuracy trade-off of Sparse-Aware Evaluation, described in Section~\ref{sec:sparse_evaluation}, on all datasets. Both variants use the same sparse VCG and differ only in forecasting: Dense Evaluation processes every pair, whereas Sparse-Aware processes active pairs and assigns $-\infty$ otherwise. The comparison therefore isolates sparse forecasting rather than the larger theoretical gain over a fully dense VCG and predictor.

Sparse-Aware results use main-experiment settings for tgbn-trade, reddit, and token. On tgbn-genre, main-setting Dense Evaluation exceeds V100 memory, so both variants use the same memory-constrained configuration for controlled comparison; its additional runtime makes the reported Sparse-Aware time unrepresentative of the main experiment. Dense Evaluation still exhausts A100 memory on reddit and token under the constrained configurations tested, whereas their reported Sparse-Aware results retain main settings. Sparse-Aware takes longer per epoch on reddit than on the larger token dataset because reddit spans more query timestamps and, as shown in Figure~\ref{fig:history_length_sensitivity}, uses a substantially longer history, respectively increasing queries per epoch and forecasting cost per active pair.

On tgbn-genre, Sparse-Aware Evaluation cuts runtime by $31.3\%$ with nearly identical NDCG@10. Because the times in Table~\ref{tab:runtime_efficiency} are rounded to one decimal place, this percentage uses unrounded averages. On smaller tgbn-trade, bookkeeping exceeds forecasting savings and increases runtime; thus, sufficient pairwise computation must be avoidable. More importantly, it runs on reddit and token where Dense Evaluation is OOM. Together with Figure~\ref{fig:temporal_sparsity_scale}, these results show that skipping inactive pairs increasingly determines feasibility as scale and sparsity grow: disable it on small datasets where overhead dominates, but enable it at scale to reduce runtime or memory.

\subsection{Sensitivity to History Length}

We select dataset-specific $L$ by validation NDCG@10 among the candidates in Figure~\ref{fig:history_length_sensitivity}; test curves are post-hoc only and do not affect selection.

Optimal $L$ is dataset-dependent. tgbn-trade peaks at $L=10$ and degrades thereafter: because each step is one year, larger $L$ quickly spans decades and introduces redundant or outdated patterns that are noisier than recent observations. tgbn-genre and reddit benefit from longer histories, peaking at $L=150$ and $210$; token saturates earlier at $L=90$ before plateauing or declining. Except for trade, performance becomes relatively insensitive to small adjustments once $L$ is sufficiently large.

Consistent with Theorem~\ref{thm:history_length_expressivity}, increasing $L$ expands temporal context and the function class, but stale or noisy interactions can hurt validation generalization. We therefore select $L$ per dataset rather than universally.

\section{Conclusion}
To solve the DNAP problem, we proposed DyNA-Mixer, combining multi-scale forecasting with a Virtual Collaborative Graph to capture pair-specific temporal evolution and localized collaboration. This separation lets the temporal and structural components specialize in complementary signals. Gated residual fusion integrates query-specific context with pair history, while Sparse-Aware Evaluation scales forecasting over the complete candidate space by processing only active pairs. 
DyNA-Mixer achieves state-of-the-art NDCG@10 on all four TGB benchmarks, with low seed variation and its largest relative gain on the largest dataset. Ablations confirm that VCG context complements temporal history, while sensitivity analysis supports dataset-specific history lengths. Sparse-Aware forecasting preserves ranking accuracy, reduces runtime when sufficient inactive computation is skipped. Together, these results establish DyNA-Mixer as an effective and scalable framework for future-affinity ranking in continuous-time dynamic graphs.

\bibliographystyle{ACM-Reference-Format}
\bibliography{sample-base}

\newpage
\appendix

\section{Notation}
\label{app:notation}

This appendix summarizes the notation used throughout the paper. Table~\ref{tab:notation} covers the dynamic graph and DNAP problem formulation, the DyNA-Mixer architecture, the Sparse-Aware Evaluation Strategy, and the theoretical analysis.

\begin{table*}[t]
\centering
\caption{Summary of major notation used in the paper.}
\label{tab:notation}
\begin{tabular}{ll}
\toprule
\textbf{Symbol} & \textbf{Description} \\
\midrule
$\mathcal{V}_{src}$, $\mathcal{V}_{tgt}$ & Sets of source and destination nodes. \\
$N$, $C$ & Numbers of source and destination nodes, respectively. \\
$\mathcal{G}_t$ & Continuous-time dynamic graph observed up to time $t$. \\
$(u_j,v_j,\tau_j,\mathbf{e}_j)$ & Interaction event with source, destination, timestamp, and edge attributes. \\
$J(t)$ & Number of interaction events observed by time $t$. \\
$u$, $c$ & Query source node and candidate destination node. \\
$\delta$ & Length of the future prediction interval. \\
$\mathbf{y}_u(t)$ & Ground-truth future affinity vector for query source $u$. \\
$\hat{\mathbf{s}}_u(t)$ & Predicted affinity score vector for query source $u$. \\
$\hat{s}_{u,c}(t)$ & Predicted affinity score for source-destination pair $(u,c)$. \\
$t_q$ & Current label timestamp; $t_1<\cdots<t_q$ denotes the label sequence. \\
$L$ & Number of preceding label timestamps retained in the history window. \\
$H_{u,c}(t_q)$ & Affinity history of pair $(u,c)$ before $t_q$, ordered from oldest to newest. \\
$\mathcal{H}$ & Collection of pair histories $\{H_{u,c}\}$. \\
$\mathcal{Q}_t$, $B_t$ & Source nodes queried at time $t$ and their number, $B_t=|\mathcal{Q}_t|$. \\
$K$ & Ranking cutoff; the experiments follow TGB and report NDCG@10. \\
\midrule
$\mathcal{G}^{*}_t$, $\mathcal{E}^{*}_t$ & Virtual Collaborative Graph and its virtual edge set at time $t$. \\
$\mathbf{m}_k(t)$, $d_{mem}$ & Continuous-time memory of destination $k$ and its dimension. \\
$\Phi(\cdot)$ & Harmonic encoder of elapsed time. \\
$R_{v,k}(t)$ & Virtual edge weight between source $v$ and destination $k$. \\
$\bar{S}_k$ & Average temporal trend associated with destination $k$. \\
$\beta_{u,k}$ & Normalized weight for combining target trends for query source $u$. \\
$G_{u,c}$ & Graph-history sequence obtained from collaborative context. \\
$d$ & Hidden dimension of the Time-Series Predictor. \\
$\mathbf{Z}^{H}_{u,c}$, $\mathbf{Z}^{G}_{u,c}$ & Embedded self-history and graph-history in $\mathbb{R}^{L\times d}$. \\
$\mathbf{W}_g$, $\mathbf{W}_u$ & Learned projections from $\mathbb{R}^{2d}$ to $\mathbb{R}^{d}$ for gating and updating. \\
$\tilde{\mathbf{Z}}_{u,c}$ & Fused temporal representation in $\mathbb{R}^{L\times d}$. \\
$\mathcal{O}_u(t)$ & Ordered destination pairs induced by ground-truth affinities for query $u$. \\
$\ell_{ij}$, $\mathcal{L}_{\mathrm{rank}}$ & Pairwise logistic loss and NDCG-weighted ranking loss. \\
$|\Delta\mathrm{NDCG}_{ij}|$ & Absolute NDCG change caused by exchanging destinations $i$ and $j$. \\
\midrule
$\mathbb{I}[\cdot]$ & Indicator function that equals one when its condition holds and zero otherwise. \\
$\mathbf{A}_{u,c}(t_q)$ & Cached number of non-zero affinities in pair $(u,c)$'s length-$L$ history before $t_q$. \\
$z_{u,c}(t_r)$ & Indicator of whether pair $(u,c)$ has a non-zero affinity at timestamp $t_r$. \\
$\mathcal{P}^{L}_{active}(t)$ & All pairs with at least one non-zero affinity in the length-$L$ history. \\
$\mathcal{P}^{L}_{inactive}(t)$ & Complement of $\mathcal{P}^{L}_{active}(t)$ in the complete pairwise space. \\
$\mathcal{P}^{L}_{active}(t;\mathcal{Q}_t)$ & Active pairs restricted to query sources in $\mathcal{Q}_t$. \\
$\mathcal{P}^{L}_{inactive}(t;\mathcal{Q}_t)$ & Inactive pairs restricted to query sources in $\mathcal{Q}_t$. \\
$\mathcal{P}^{\infty}_{active}(t)$ & Pairs with a non-zero affinity at any observed label timestamp before $t$. \\
$\rho_L(t)$, $\rho_{\infty}(t)$ & Active-pair ratios under finite and infinite histories, respectively. \\
$\mathcal{T}(L,d)$ & Cost of applying the Time-Series Predictor to one length-$L$ sequence. \\
$X(L,d)$ & Expensive gated-fusion and forecasting cost for one pair. \\
$M_t$, $r_t$ & Number and ratio of active queried pairs at time $t$. \\
$\bar{r}$ & Query-weighted active-pair ratio across all timestamps. \\
$\mathcal{F}_L$ & Function class representable using histories of length $L$. \\
\bottomrule
\end{tabular}
\end{table*}

\section{Theoretical Analysis}
\label{app:theoretical_analysis}

\subsection{Infinite-History Sparsity Bound}

Cumulative temporal sparsity is measured using the complete observed history in Section~\ref{sec:empirical_sparsity}. We now show that this measurement is a conservative upper bound on the active-pair ratio for every finite history window.

Let $t_1<\cdots<t_q$ denote the label timestamps observed before the current prediction target. Under a finite history length $L$, define

\begin{equation}
\mathcal{P}^{L}_{active}(t_q)
=\left\{(u,c)\,\middle|\,
\exists r\in\{\max(1,q-L),\ldots,q-1\},
\mathbf{y}_u(t_r)[c]>0
\right\},
\end{equation}

where $(u,c)\in\mathcal{V}_{src}\times\mathcal{V}_{tgt}$. The corresponding infinite-history set is

\begin{equation}
\mathcal{P}^{\infty}_{active}(t_q)
=\left\{(u,c)\,\middle|\,
\exists r<q,\ \mathbf{y}_u(t_r)[c]>0
\right\}.
\end{equation}

Their active-pair ratios are

\begin{equation}
\rho_L(t_q)=\frac{|\mathcal{P}^{L}_{active}(t_q)|}{NC},
\qquad
\rho_{\infty}(t_q)=\frac{|\mathcal{P}^{\infty}_{active}(t_q)|}{NC}.
\end{equation}

\begin{theorem}[Infinite-history upper bound]
\label{thm:infinite_history_bound}
For every finite $L$ and label timestamp $t_q$,

\begin{equation}
\rho_L(t_q)\leq\rho_{\infty}(t_q).
\end{equation}

When $q>L$, the inequality is strict whenever at least one pair has a non-zero affinity before $t_{q-L}$ but no non-zero affinity within the current length-$L$ window.
\end{theorem}

\begin{proof}
The finite window contains only label timestamps from $t_{\max(1,q-L)}$ to $t_{q-1}$, all of which also belong to the complete history before $t_q$. Therefore,

\begin{equation}
\mathcal{P}^{L}_{active}(t_q)
\subseteq
\mathcal{P}^{\infty}_{active}(t_q).
\end{equation}

Taking cardinalities and dividing by the positive constant $NC$ gives $\rho_L(t_q)\leq\rho_{\infty}(t_q)$. If a pair was active only before the current finite window, it belongs to $\mathcal{P}^{\infty}_{active}(t_q)$ but not to $\mathcal{P}^{L}_{active}(t_q)$, making the set inclusion and ratio inequality strict.
\end{proof}

Equivalently, the inactive-pair ratio satisfies $1-\rho_L(t_q)\geq1-\rho_{\infty}(t_q)$. Thus, the sparsity shown in Figure~\ref{fig:temporal_sparsity_scale} does not overstate the sparsity available to any finite-history model.

\subsection{Sparse Approximation and Ranking Preservation}

Sparse-Aware Evaluation bypasses VCG aggregation, gated fusion, and temporal forecasting when $H_{u,c}=\mathbf{0}$, assigning the corresponding pair a score of $-\infty$. This differs from dense evaluation, under which an inactive pair could still receive a finite score from collaborative context alone. The strategy is therefore an approximation whose effect depends on whether inactive pairs would enter the reported ranking.

Fix a query source $u$ and time $t$, and suppose that at least $K$ of its candidate destinations are active. Let $a_K(u,t)$ denote the $K$-th highest score assigned by dense evaluation to these active destinations, and let $b(u,t)$ denote the highest dense-evaluation score among the inactive destinations. Sparse-Aware Evaluation preserves the Top-$K$ ranking whenever

\begin{equation}
a_K(u,t)>b(u,t).
\end{equation}

Under this condition, no inactive destination can enter the dense Top-$K$ result. Sparse-Aware Evaluation leaves the scores of active pairs unchanged and suppresses only inactive pairs; therefore, it produces the same Top-$K$ ranking and NDCG@$K$ as dense evaluation.

The finite-history update also prevents errors on newly relevant pairs from persisting. Consider a pair that is inactive at $t_q$ but has a positive ground-truth affinity in the current prediction target. Once $\mathbf{y}_u(t_q)$ is observed, this positive value is appended to its history. For every $L\geq1$,

\begin{equation}
H_{u,c}(t_q)=\mathbf{0}
\ \land\ 
\mathbf{y}_u(t_q)[c]>0
\ \Longrightarrow\
(u,c)\in\mathcal{P}^{L}_{active}(t_{q+1}).
\end{equation}

Such a pair may be suppressed incorrectly at $t_q$, but it is processed by the complete model from the next label timestamp onward. The approximation error associated with a newly relevant pair is therefore transient rather than persistent. By contrast, an inactive pair whose current ground-truth affinity remains zero does not become active; ranking it highly is a false positive, which the LambdaRank loss penalizes when it displaces a relevant destination. These two mechanisms limit sustained ranking errors from inactive pairs. Consistently, Sparse-Aware Evaluation changes test NDCG@10 by only $0.00010$, as reported in Table~\ref{tab:runtime_efficiency}.

\subsection{Complexity and Scalability}

The per-timestamp complexities of dense and sparse evaluation are given in Section~\ref{sec:sparse_evaluation}. We now derive how the active-pair ratio determines their relative costs. Define the expensive per-pair computation as

\begin{equation}
X(L,d)=Ld^2+\mathcal{T}(L,d).
\end{equation}

Let $M_t$ be the number of active queried pairs. The corresponding active-pair ratio is

\begin{equation}
r_t=\frac{M_t}{B_tC},
\qquad
M_t=\left|\mathcal{P}^{L}_{active}(t;\mathcal{Q}_t)\right|.
\end{equation}

Ignoring constant factors, the dense-evaluation cost is

\begin{equation}
D_t=B_tC\,X(L,d).
\end{equation}

The sparse-evaluation cost is

\begin{equation}
S_t=B_tC+M_tX(L,d)
=B_tC\left(1+r_tX(L,d)\right).
\end{equation}

Consequently, the idealized speedup ratio is

\begin{equation}
\frac{D_t}{S_t}
=\frac{X(L,d)}{1+r_tX(L,d)}.
\end{equation}

As the per-pair predictor becomes more expensive, the limiting ratio is

\begin{equation}
\lim_{X(L,d)\rightarrow\infty}\frac{D_t}{S_t}
=\frac{1}{r_t}.
\end{equation}

Thus, the reciprocal active-pair ratio determines the asymptotic benefit of skipping deep computation. The additive $B_tC$ term also shows why sparsity cannot eliminate all work: the DNAP problem still requires identifying the active support and returning $C$ scores for each query source.

A similar result holds for intermediate memory. Dense evaluation materializes temporal representations for all pairs, whereas sparse evaluation retains dense output scores but materializes length-$L$ hidden states only for active pairs. Their idealized memory ratio is

\begin{equation}
\frac{B_tCLd}{B_tC+M_tLd}
=\frac{Ld}{1+r_tLd}.
\end{equation}

This ratio also approaches $1/r_t$ as $Ld$ grows.

The count cache provides an additional saving during support identification. Recomputing activity from the stored histories would inspect $B_tCL$ values at each query timestamp. Incrementally removing the expired value and adding the newest value requires only $B_tC$ updates, reducing support maintenance by a factor of $L$. This cache changes neither the affinity histories nor the prediction rule; it only avoids repeatedly scanning the same window.

Finally, the query-weighted active ratio across all timestamps is

\begin{equation}
\bar{r}
=\frac{\sum_t M_t}{C\sum_t B_t}.
\end{equation}

When $L$ and $d$ are fixed, the cumulative dense-to-sparse cost ratio has the same form:

\begin{equation}
\frac{X(L,d)}{1+\bar{r}X(L,d)}.
\end{equation}

If every source is queried at time $t$, then $B_t=N$ and $r_t=\rho_L(t)\leq\rho_{\infty}(t)$ by Theorem~\ref{thm:infinite_history_bound}. For a query subset, its active ratio need not equal the global ratio in Figure~\ref{fig:temporal_sparsity_scale}; the runtime experiment therefore measures the realized speedup, including query composition, fixed costs, and sparse-processing overheads.

\subsection{History Length and Model Expressivity}

The history length $L$ determines how many previously observed affinity values are available to the predictor. We compare the corresponding function classes under the assumption that additional historical positions can be masked, allowing a longer-history model to ignore them. Let $\mathcal{F}_L$ denote the prediction functions representable from a length-$L$ history under this model family.

\begin{theorem}[Monotonic expressivity with history length]
\label{thm:history_length_expressivity}
For $L_1<L_2$,

\begin{equation}
\mathcal{F}_{L_1}\subseteq\mathcal{F}_{L_2}.
\end{equation}

If the predictor can use at least one of the additional $L_2-L_1$ historical positions, then the containment is strict.
\end{theorem}

\begin{proof}
At query timestamp $t_q$, the most recent $L_1$ entries of a length-$L_2$ history are

\begin{equation}
H^{(L_1)}_{u,c}(t_q)
=\left[
\mathbf{y}_u(t_{q-L_1})[c],\ldots,
\mathbf{y}_u(t_{q-1})[c]
\right].
\end{equation}

For any $f\in\mathcal{F}_{L_1}$, a length-$L_2$ model can mask its first $L_2-L_1$ entries and apply the same computation to this suffix. Hence every function in $\mathcal{F}_{L_1}$ is also in $\mathcal{F}_{L_2}$. If the predictor can assign non-zero influence to an additional position, it can represent a function whose output changes when that position changes while the most recent $L_1$ values remain fixed. No length-$L_1$ model can represent such a function, so the containment is strict.
\end{proof}

Increasing $L$ therefore expands the temporal information that the model can use, but greater expressivity does not guarantee better generalization. Longer histories may contain stale or noisy interactions and can increase overfitting risk. We consequently select $L$ using validation performance, as reported in the sensitivity experiment.

\section{Trainable Parameter Count}
\label{app:model_size}

The number of trainable parameters under different history lengths is reported in Tables~\ref{tab:model_size_trade}--\ref{tab:model_size_token}. We report only trainable parameters here, rather than runtime memory consumption, because dataset-dependent historical states and temporary tensors are not part of the learnable model capacity. The parameter count is affected by the history length $L$: longer histories increase the input dimensionality of the temporal predictor and also slightly enlarge the VCG relation module, whose virtual-edge inference uses historical affinity sequences as part of its input. The dagger symbol ($^\dagger$) marks the history length selected for the main experiments.

\begin{table*}[t]
\centering
\caption{Trainable parameter count of DyNA-Mixer on tgbn-trade under different history lengths. The dagger symbol ($^\dagger$) marks the history length selected for the main experiments.}
\label{tab:model_size_trade}
\resizebox{0.8\textwidth}{!}{
\begin{tabular}{llrrr}
\toprule
\textbf{Dataset} & \textbf{History Length $L$} & \textbf{TSP Params} & \textbf{VCG Params} & \textbf{Total Params} \\
\midrule
tgbn-trade & 3  & 21,475 & 122,737 & 144,212 \\
tgbn-trade & $10^\dagger$ & 22,085 & 122,961 & 145,046 \\
tgbn-trade & 30 & 26,895 & 123,601 & 150,496 \\
tgbn-trade & 60 & 42,735 & 124,561 & 167,296 \\
\bottomrule
\end{tabular}
}
\end{table*}

\begin{table*}[t]
\centering
\caption{Trainable parameter count of DyNA-Mixer on tgbn-genre under different history lengths. The dagger symbol ($^\dagger$) marks the history length selected for the main experiments.}
\label{tab:model_size_genre}
\resizebox{0.8\textwidth}{!}{
\begin{tabular}{llrrr}
\toprule
\textbf{Dataset} & \textbf{History Length $L$} & \textbf{TSP Params} & \textbf{VCG Params} & \textbf{Total Params} \\
\midrule
tgbn-genre & 3   & 23,539  & 122,737 & 146,276 \\
tgbn-genre & 10  & 24,149  & 122,961 & 147,110 \\
tgbn-genre & 30  & 28,959  & 123,601 & 152,560 \\
tgbn-genre & 60  & 44,799  & 124,561 & 169,360 \\
tgbn-genre & 90  & 70,989  & 125,521 & 196,510 \\
tgbn-genre & 120 & 107,529 & 126,481 & 234,010 \\
tgbn-genre & $150^\dagger$ & 154,419 & 127,441 & 281,860 \\
tgbn-genre & 180 & 211,659 & 128,401 & 340,060 \\
tgbn-genre & 210 & 279,249 & 129,361 & 408,610 \\
\bottomrule
\end{tabular}
}
\end{table*}

\begin{table*}[t]
\centering
\caption{Trainable parameter count of DyNA-Mixer on tgbn-reddit under different history lengths. The dagger symbol ($^\dagger$) marks the history length selected for the main experiments.}
\label{tab:model_size_reddit}
\resizebox{0.8\textwidth}{!}{
\begin{tabular}{llrrr}
\toprule
\textbf{Dataset} & \textbf{History Length $L$} & \textbf{TSP Params} & \textbf{VCG Params} & \textbf{Total Params} \\
\midrule
tgbn-reddit & 3   & 25,019  & 122,737 & 147,756 \\
tgbn-reddit & 10  & 25,629  & 122,961 & 148,590 \\
tgbn-reddit & 30  & 30,439  & 123,601 & 154,040 \\
tgbn-reddit & 60  & 46,279  & 124,561 & 170,840 \\
tgbn-reddit & 90  & 72,469  & 125,521 & 197,990 \\
tgbn-reddit & 120 & 109,009 & 126,481 & 235,490 \\
tgbn-reddit & 150 & 155,899 & 127,441 & 283,340 \\
tgbn-reddit & 180 & 213,139 & 128,401 & 341,540 \\
tgbn-reddit & $210^\dagger$ & 280,729 & 129,361 & 410,090 \\
tgbn-reddit & 240 & 358,669 & 130,321 & 488,990 \\
\bottomrule
\end{tabular}
}
\end{table*}

\begin{table*}[t]
\centering
\caption{Trainable parameter count of DyNA-Mixer on tgbn-token under different history lengths. The dagger symbol ($^\dagger$) marks the history length selected for the main experiments.}
\label{tab:model_size_token}
\resizebox{0.8\textwidth}{!}{
\begin{tabular}{llrrr}
\toprule
\textbf{Dataset} & \textbf{History Length $L$} & \textbf{TSP Params} & \textbf{VCG Params} & \textbf{Total Params} \\
\midrule
tgbn-token & 3   & 27,443  & 122,865 & 150,308 \\
tgbn-token & 10  & 28,053  & 123,089 & 151,142 \\
tgbn-token & 30  & 32,863  & 123,729 & 156,592 \\
tgbn-token & 60  & 48,703  & 124,689 & 173,392 \\
tgbn-token & $90^\dagger$ & 74,893  & 125,649 & 200,542 \\
tgbn-token & 120 & 111,433 & 126,609 & 238,042 \\
tgbn-token & 150 & 158,323 & 127,569 & 285,892 \\
\bottomrule
\end{tabular}
}
\end{table*}

The results show that DyNA-Mixer remains lightweight in terms of learnable capacity. Even for the largest selected setting, tgbn-reddit with $L=210$, the full model contains only 410,090 trainable parameters. Increasing $L$ mainly affects the Time-Series Predictor, while the VCG module grows more slowly because its shared relation network is reused across source-destination pairs. This supports that the main computational challenge of dynamic node affinity prediction comes from dense temporal evaluation over large candidate spaces, rather than from an excessively large number of trainable parameters.


\section{Training Dynamics}
\label{app:training_dynamics}

\begin{figure*}[t]
  \centering
  \includegraphics[width=0.90\textwidth]{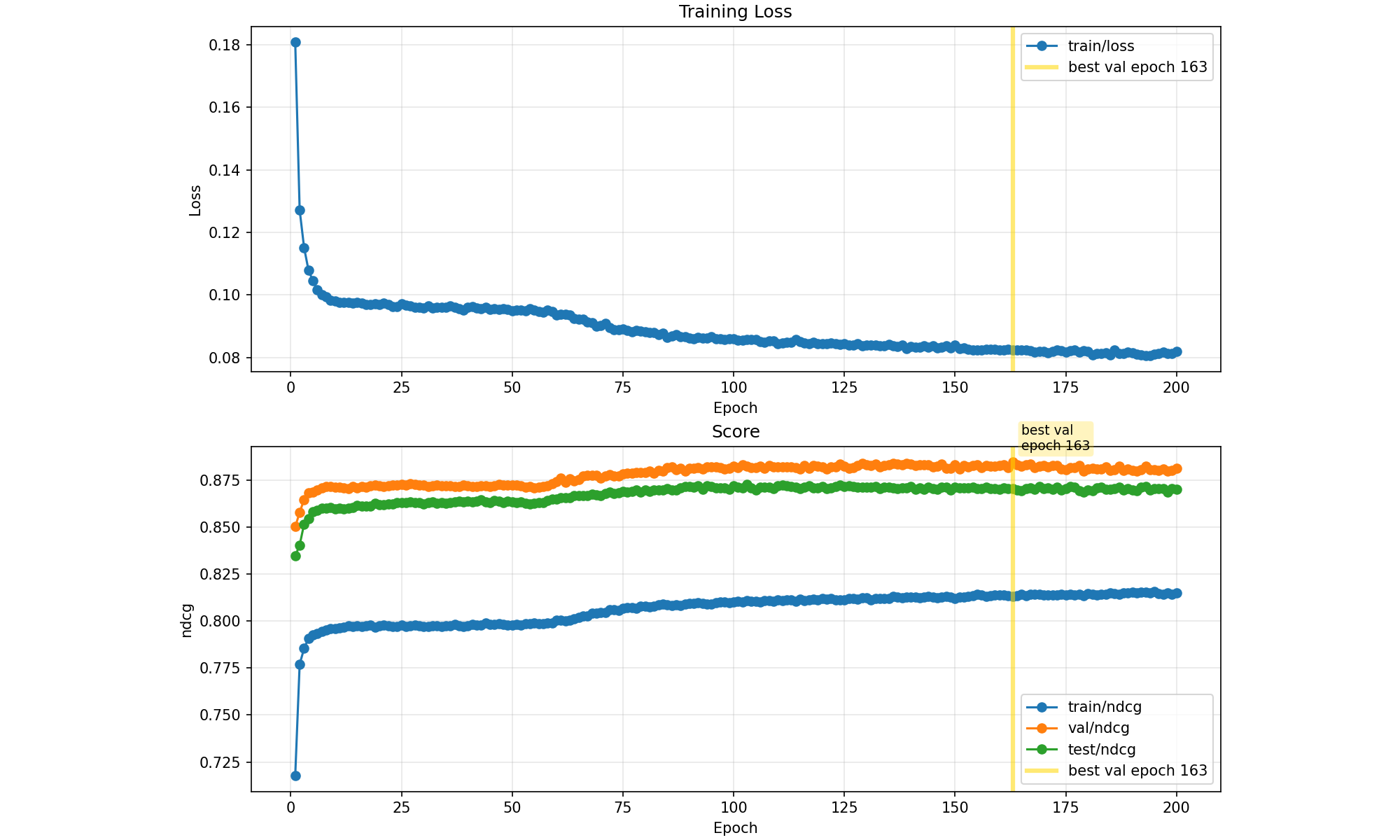}
  \caption{Training dynamics of DyNA-Mixer on tgbn-trade under the selected history length $L=10$.}
  \label{fig:training_trade}
\end{figure*}

\begin{figure*}[t]
  \centering
  \includegraphics[width=0.90\textwidth]{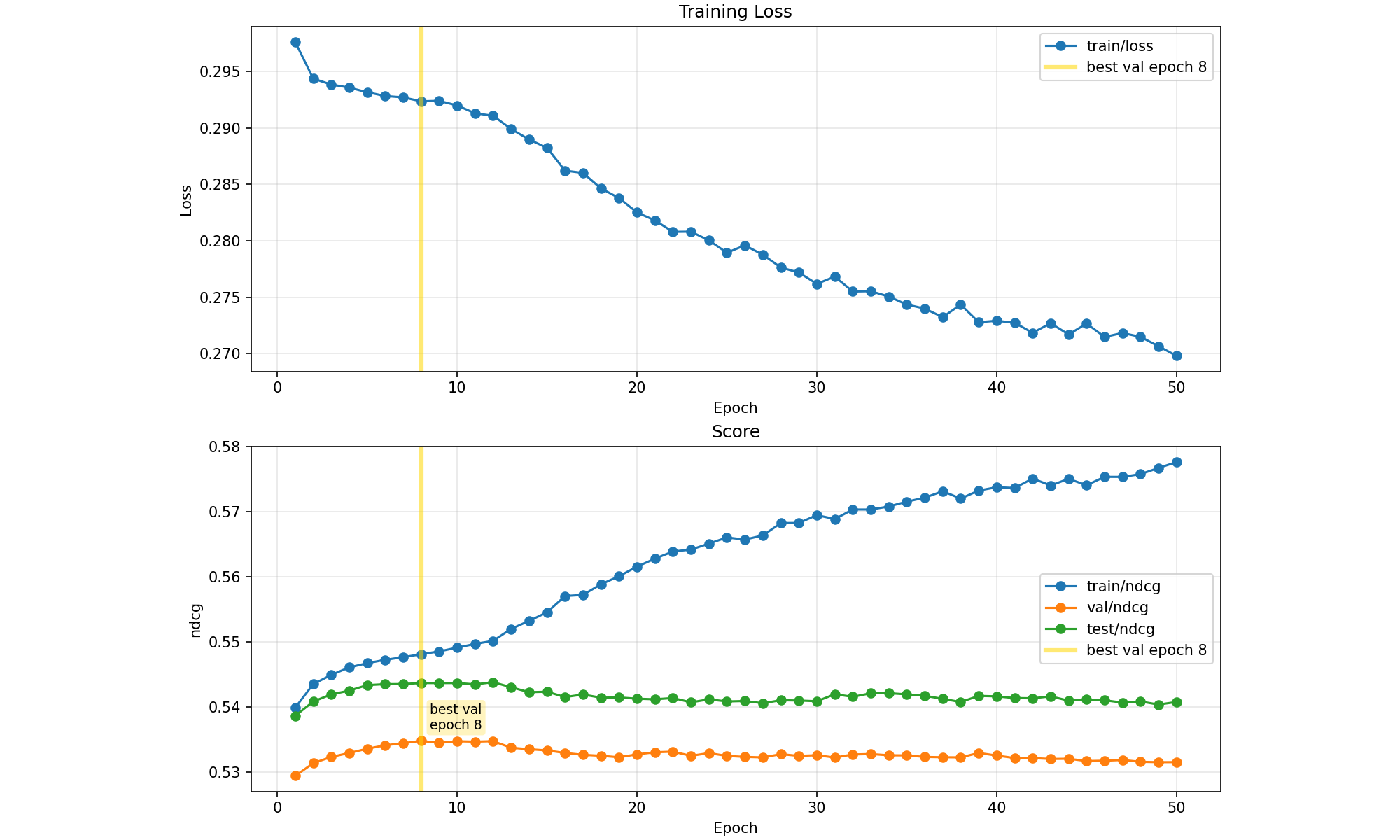}
  \caption{Training dynamics of DyNA-Mixer on tgbn-genre under the selected history length $L=150$.}
  \label{fig:training_genre}
\end{figure*}

\begin{figure*}[t]
  \centering
  \includegraphics[width=0.90\textwidth]{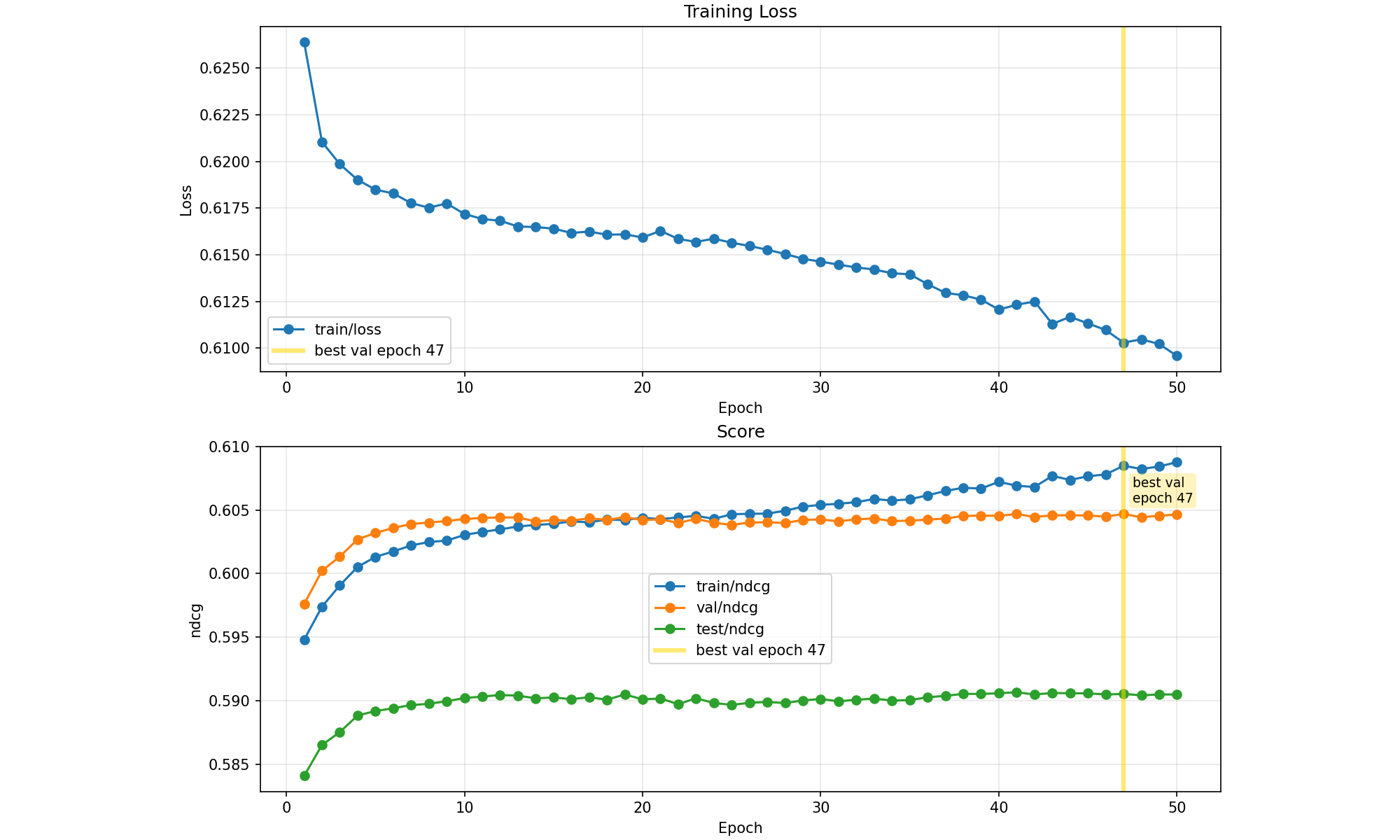}
  \caption{Training dynamics of DyNA-Mixer on tgbn-reddit under the selected history length $L=210$.}
  \label{fig:training_reddit}
\end{figure*}

\begin{figure*}[t]
  \centering
  \includegraphics[width=0.90\textwidth]{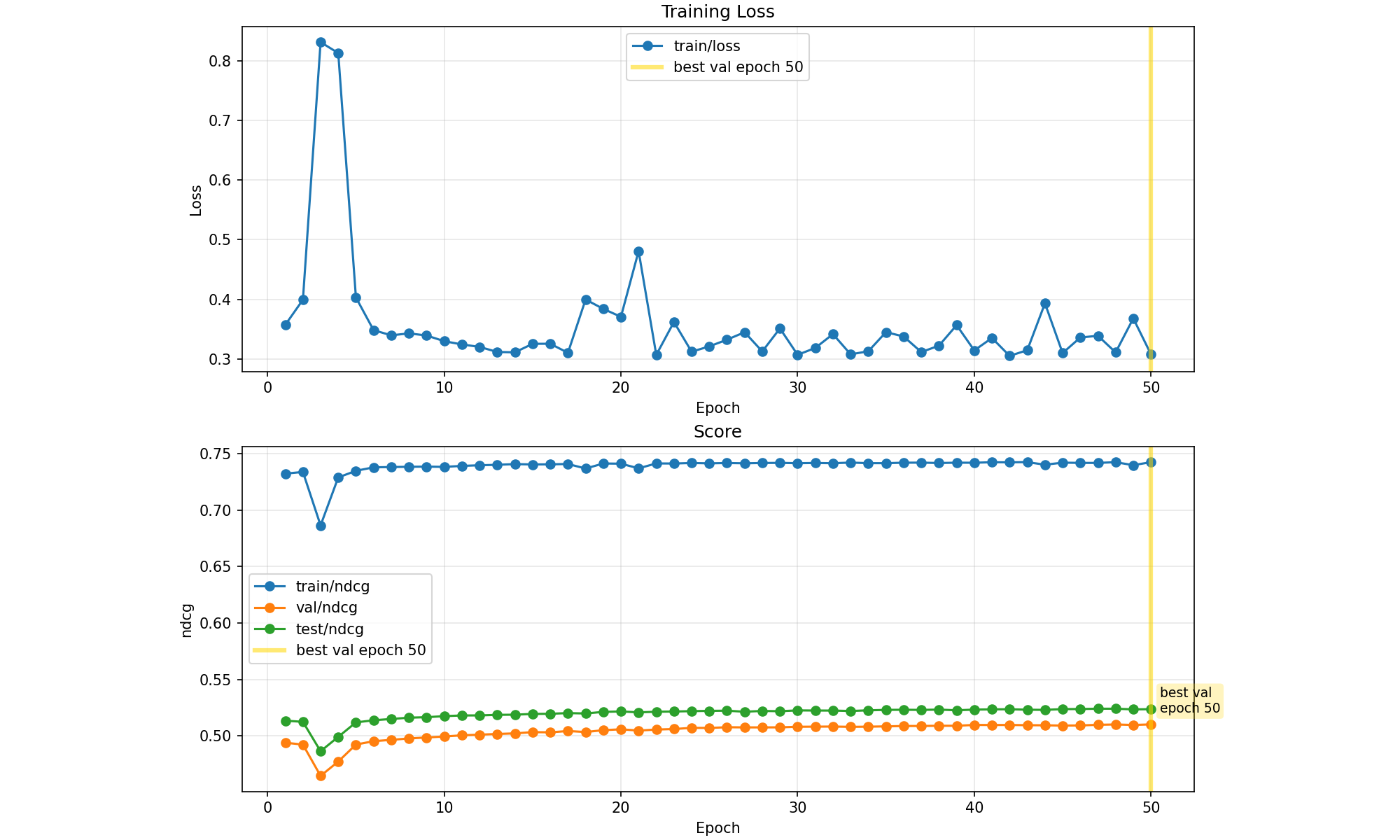}
  \caption{Training dynamics of DyNA-Mixer on tgbn-token under the selected history length $L=90$.}
  \label{fig:training_token}
\end{figure*}

The training dynamics of DyNA-Mixer under the selected history length for each dataset are presented in Figures~\ref{fig:training_trade}--\ref{fig:training_token}. Each figure reports the training loss and NDCG@10 on the training, validation, and test splits. The vertical line marks the epoch with the best validation performance, which is used for model selection.

\end{document}